\documentclass[runningheads,a4paper]{llncs}

\usepackage{amssymb}
\usepackage{graphicx}

\usepackage{url}
\urldef{\mailsa}\path|{a.kota.gopalakrishna, t.ozcelebi, a.liotta, j.j.lukkien}@tue.nl|
\newcommand{\keywords}[1]{\par\addvspace\baselineskip
\noindent\keywordname\enspace\ignorespaces#1}

\begin{document}

\title{Relevance As a Metric for Evaluating Machine Learning Algorithms}

\titlerunning{Relevance As a Metric for Evaluating Machine Learning Algorithms}

\author{Aravind Kota Gopalakrishna, Tanir Ozcelebi, Antonio Liotta, and Johan J. Lukkien}

\authorrunning{A. K. Gopalakrishna et. al.}

\institute{System Architecture and Networking (SAN),\\
Department of Mathematics and Computer Science, \\
\and
Electro-Optical Communications, \\
Department of Electrical Engineering,\\
Eindhoven University of Technology, The Netherlands\\
\mailsa\\
}

\toctitle{Relevance As a Metric for Evaluating Machine Learning Algorithms}
\tocauthor{Aravind Kota Gopalakrishna, Tanir Ozcelebi, Antonio Liotta, and Johan J. Lukkien}
\maketitle

\begin{abstract}
In machine learning, the choice of a learning algorithm that is suitable for the application domain is critical. The performance metric used to compare different algorithms must also reflect the concerns of users in the application domain under consideration. In this work, we propose a novel probability-based performance metric called \emph{Relevance Score} for evaluating supervised learning algorithms. We evaluate the proposed metric through empirical analysis on a dataset gathered from an intelligent lighting pilot installation. In comparison to the commonly used Classification Accuracy metric, the Relevance Score proves to be more appropriate for a certain class of applications.
\end{abstract}

\section{Introduction}\label{INTRO}
One of the general goals of a machine learning (ML) algorithm is to capture complex relationships and patterns within a dataset and to apply such knowledge on new data that is believed to have the same (or a similar) pattern. The success criteria for the ML algorithm depend on many factors, most importantly, on the user concerns that are specific to the application domain.

Consider a car navigation system that utilizes a specific ML algorithm to select the best route from a source location to a destination. There are many ways to measure the performance of this algorithm objectively. Useful metrics can be, for instance, the actual time to reach the destination, the amount of gas consumption and the amount of tolls paid. However, it is often the case that a combination of these metrics cannot directly be mapped to the real satisfaction of the driver from the navigation experience, whereas this is what matters in the end. Assume that there are only two home-office routes possible and the driver \emph{generally} likes to take the first route, which is shorter, faster and safer. On the other hand, \emph{sometimes}, the driver likes to take the second route that goes along the sea shore, especially when it is sunny and the driver is in the mood. The ML algorithm can be expected to learn to pick the desired route among the two only if weather is a parameter that can be and is monitored by the navigation device. Monitoring the mood of a person is very difficult nonetheless. Alternatively, the ML algorithm will simply start looking for patterns within other data that is available to it, e.g. the time of the day, which may or may not have any actual impact. In this case, the desired route may also depend on those parameters, i.e. features, which are relevant yet not considered by the navigation device.

It is a difficult task to identify \emph{all} the features that are relevant for the ML algorithm employed by an application. The features that are relevant for the application define a \emph{context}. Even when the context is known, it may be technically very challenging or very intrusive to monitor some features especially when users are involved, e.g. the driver's mood in the previous example. The subset of relevant features that are monitored define an \emph{observed context}. Depending on the application under consideration and the observed context, sometimes evaluating an output, i.e. a prediction made by the ML algorithm, as either \emph{right} or \emph{wrong} might not be sufficient as there are gray areas in between. For example, the intelligent environment described in~\cite{AG} offers lighting services (light settings) depending on several contextual factors such as user identity, user activity and time of the day. When the dataset collected from this intelligent environment is examined, there are two important remarks to be made. The first remark is that the desired light settings are not fixed for a given instance of observed context. Since perception of lighting also depends on subjective factors that cannot be monitored, even the same user, the same activity and the same time of the day may result in different desired light settings. The second remark is that, for a given observed context, there are acceptable (desired at least once) and unacceptable light settings (never desired).

This means that the overall performance of the ML algorithm may be hindered by the fact that the observed context is not entirely representative of the actual context of an application. When this is the case, it is a challenge to evaluate and compare the performances of different ML algorithms in the application domain. In the literature, different evaluation metrics assess different characteristics of ML algorithms~\cite{MS}. Classification problems in ML are broadly categorized into multi-class classification and multi-label classification problems. Multi-class classification algorithms~\cite{MA} are for those categories of problems where, for a given input instance, there is exactly one correct output class, which is selected out of several classes. Examples of multi-class classification include image recognition and diagnosis~\cite{IW}. The commonly used evaluation metric for multi-class classification problems are accuracy, precision and recall. Multi-label classification algorithms~\cite{GT} address that category of problems where multiple output classes must be selected for each input instance. Examples of multi-label classification include text~\cite{AK} and music categorization~\cite{KT}. The commonly used evaluation metrics for multi-label classification problems are hamming-loss, precision and recall.

As illustrated by the examples of car navigation and intelligent lighting, there also exists another class of applications. In these applications, the ML prediction model has to select a unique output class for a given input instance (as in multi-class classification), and the output for a given input instance may fall into multiple acceptable output classes (as in multi-label classification). An interesting property of such an application is randomness in the output, i.e. there is no single acceptable outcome for a given observed context but there is a statistical regularity associated with different possible and acceptable outcomes. In such cases, even though the prediction models that are used to solve multi-class classification problems suits the need, the evaluation metrics such as accuracy, precision and recall are not suitable. For instance, the use of accuracy as a metric would lead to a score of zero for a misclassified sample, which should not be the case.  The bad choice of performance metrics would produce results that are meaningless to achieve the design goal of an application~\cite{NJ}~\cite{RJ}.

In this work, we devise a new evaluation scheme to evaluate the performance of supervised learning algorithms for this class of applications and propose a new evaluation metric called \emph{Relevance Score (RS)}. We consider an intelligent lighting scenario to describe and formulate the problem. \emph{Relevance Score} is defined as a percentage measure of how much relevant an output or outcome is, as predicted by the prediction model for a given observed context. The score is based on the probabilistic distances among the different outcomes, i.e. predicted and actual outcome. We use this metric to evaluate several rule based prediction models and then compare it with the commonly used metric \emph{Prediction Accuracy or Classification Accuracy (CA)} to validate the significance of the proposed metric.

The paper is organized as follows. Section~\ref{PD} discusses the drawbacks of using the commonly used \emph{CA} metric with an example of intelligent lighting application and identifies improvement directions. Section~\ref{PM} presents the proposed relevance metric. Section~\ref{ER} provides an evaluation of the proposed metric. Finally, Section~\ref{CONC} concludes the paper.

\section{Problem Definition}\label{PD}
In this section, we discuss the drawbacks of \emph{Classification Accuracy (CA)} as a metric for evaluating supervised learning algorithms, which motivates the need for a new metric using an intelligent lighting application scenario.

\begin{figure}[!t]
    \centering
        \includegraphics[height=8cm]{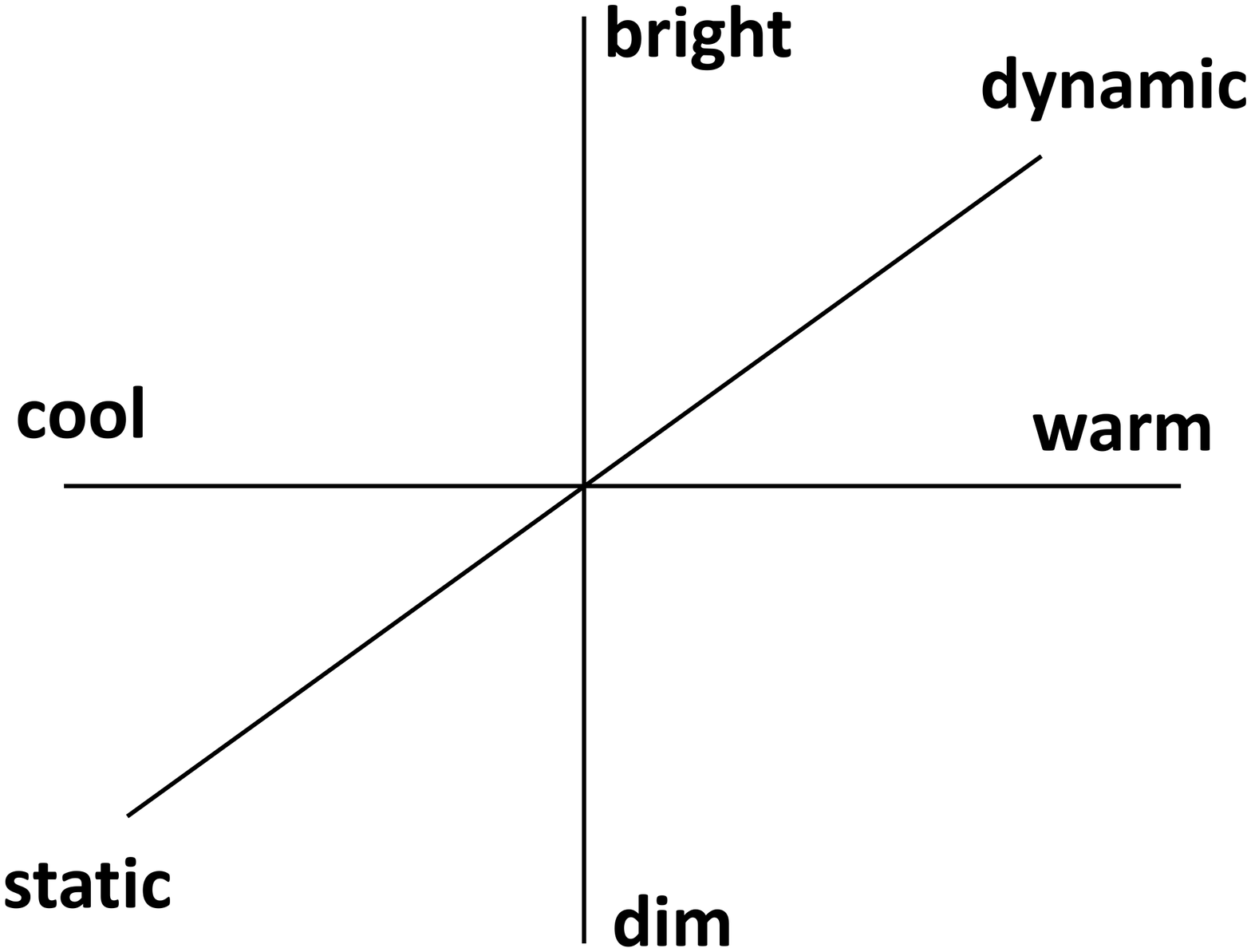}
    \caption{Possible Output Light Combinations}\label{fig1}
\end{figure}

Let $x_i\epsilon X$ denote the $i^{th}$ sample in a dataset, where $x_i=[x_{i1},x_{i2},\ldots,x_{in}]$ and $X$ represents an $n$-dimensional input feature space given by $X=X_1\times X_2\times X_3\times \ldots \times X_n$. Let us denote the output class label as $y$ where $y\epsilon Y = \{y_1,y_2,\ldots,y_K\}$  and $K$ denotes the number of possible outcomes. In the intelligent lighting application under consideration~\cite{AG}, there are 8 possible presets for output lighting conditions (K=8) in a breakout area, as shown in the three dimensional space in Fig.~\ref{fig1}. Each octant represents a possible light combination (e.g. static-warm-dim).

 A user is provided with a lighting condition based on six input features $(n=6)$ namely user-identity, type of activity, area of activity, number of users, time of the day and the external light influence.

In~\cite{AG}, the supervised learning approach is explored where a considered classification model is trained on the breakout dataset to predict a unique light output $y_i\epsilon Y = \{y_1,y_2,\ldots,y_8\}$ for a given input $x_i=[x_{i1},x_{i2},\ldots,x_{i6}]$. The success of an intelligent lighting application depends on selecting the best suitable prediction model that selects the desired lighting condition for a given observed context. The performance of the selected models is evaluated using \emph{CA} as a metric.

Let us assume that a classification model \emph{h} is trained using the breakout dataset. The breakout dataset contains a sample \emph{x} 10 times and the output light condition selected by a user is light A \emph{(LA)} four times, light B \emph{(LB)} four times and light C \emph{(LC)} two times. This means that the user for a given input instance \emph{x} has selected \emph{LA}, \emph{LB} and \emph{LC} with probabilities 0.4, 0.4 and 0.2 respectively, i.e. a sample \emph{x} has multiple output classes. Assume that the classifier \emph{h} selects \emph{LA} based on what it has learnt from the training data, whereas the user actually desires \emph{LB} as the lighting condition. Here the classifier \emph{h} is not right with the prediction, but not entirely wrong either. If \emph{CA} is used as a metric, then the selection made by the classifier \emph{h} would be assessed as completely wrong. From the application point-of-view the \emph{CA} metric is not very representative as it is important to measure how relevant the lighting condition is for a given observed context, rather than to measure how accurate it is. It can never be 100\% accurate as the user is not consistent in choosing the desired lighting condition, based on the observed context.

From the example above it is evident that CA is not a relevant metric for similar applications as it fails to capture the degree of relevance when there is a ``non-ideal" prediction. Hence, it is necessary to have an evaluation mechanism that is able to compute the distance between actual and predicted outcome when they are not equal. An example of such a metric for a given classifier is shown in Table~\ref{table1}.

\begin{table}[!t]
\renewcommand{\arraystretch}{1.3}
\caption{Sample comparison of metrics}\label{table1}
\begin{center}
     \begin{tabular}{ | c | c | c | c |}
     \hline
     Actual Output & Predicted Output & CA Metric & Alternative Metric \\
     \hline
     $LA$ & $LA$ & 100 & Good match \\
           &       &     & (100) \\ \hline
     $LB$ & $LC$ & 0 & Decent match \\
           &       &     & (~75) \\ \hline
     $LC$ & $LA$ & 0 & Irrelevant match \\
           &       &     & (0) \\ \hline
     $LA$ & $LB$ & 0 & Relevant match  \\
           &       &     & (~50) \\ \hline
     $LB$ & $LB$ & 100 &  Good match   \\
           &       &     & (100) \\ \hline
     &      & 200/5 = 40\% & 325/5 = 65\% \\
     \hline
     \end{tabular}
 \end{center}
 \end{table}

 As shown in Table~\ref{table1}, by virtue of the problem, \emph{CA} gives a low accuracy of 40\%, whereas the user is given relevant outputs 80\% of the time and irrelevant outputs only 20\% of the time according to the \emph{Alternative Metric}, resulting in an average score of 65\%. Here the \emph{ Alternative Metric} computes the relevance of the outcome i.e. how good is the output selection made by a prediction model through the data statistics available from the observed environment, thereby making it more suitable than the \emph{CA} metric.

Formally, we propose an evaluation metric named as \emph{Relevance Score}, which is more suitable for the third class of applications mentioned in Section~\ref{INTRO} than the commonly used CA metric. In particular, we seek to find a performance function $f:(X,Y)\rightarrow R^{100}_{+}$ where  $R^{100}_{+}=\{RS \epsilon R | 0\leq RS\leq 100\}$.

\section{Proposed Metric}\label{PM}

In this section, we propose a new mechanism to evaluate supervised prediction models used in the presented class of applications, e.g. in intelligent lighting. Here, we formulate a function that quantifies the relevance of the predicted outcome when there is a mismatch between the predicted and actual outcome.

The well known CA metric is computed as an average of the sum of individual accuracies for $k$ test set samples as in equation~\ref{Eqn1}.

\begin{equation}\label{Eqn1}
    CA=\frac{1}{k}\sum_{i=1}^{k}Acc_i
\end{equation}

We inherit the same mechanism to compute the \emph{Relevance Score }on a test set for a given prediction model with a minor modification where the accuracy \emph{Acc} for a sample is replaced by \emph{Score} as in equation~\ref{Eqn2}.

\begin{figure}[!t]
    \centering
        \includegraphics[height=8cm]{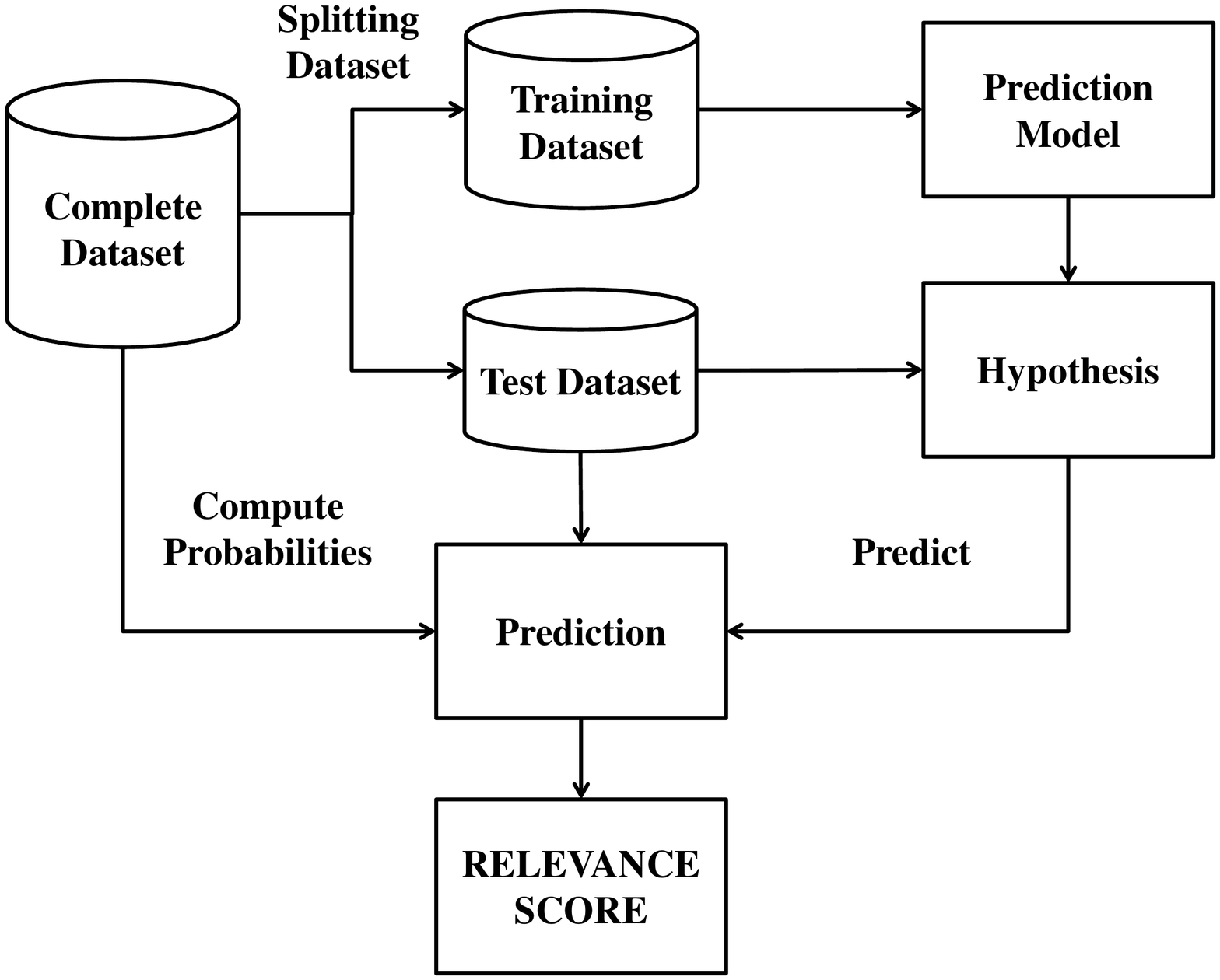}
    \caption{Relevance Score computation procedure}\label{fig2}
\end{figure}

\begin{equation}\label{Eqn2}
    RS=\frac{1}{k}\sum_{i=1}^{k}Score_i
\end{equation}

We divide  \emph{RS} computation into two phases: 1. Probability Computation Phase, and 2. Evaluation Phase. The entire process of computing \emph{RS} is shown in Fig.~\ref{fig2}.

\subsection{Probability Computation Phase}
 An \emph{ErrScore} is a real number that is calculated when there is an error in prediction by the prediction model. It is computed based on five parameters, the predicted outcome $(O_P)$, the actual outcome $(O_A)$, the probability of occurrence of the predicted outcome $(P(O_P))$, the probability of occurrence of the actual outcome $(P(O_A))$ and the probability of the most frequently selected outcome $(P(O_H))$. These probabilities are the values that are computed from the entire dataset. The probability values are computed for the different output class labels $(y \epsilon Y)$ for every input instance $(x_i)$ in the breakout dataset. Formally,  compute $P(y|x_i)$ for $i=1,2,3,...,m$ where $m$ is the number of samples in the entire dataset. In this phase, the values of the most probable feature that is responsible for the randomness in output (e.g. user identity feature in intelligent lighting application) is removed for generalization.

\subsection{Evaluation Phase}
In this phase, the predictions on the test data are evaluated using the considered classification models. If $d$ denotes the probabilistic distances in general, then the distances between the probabilities are given by,

\begin{equation}
    d_{HP}=|P(O_H)-P(O_P)|
\end{equation}
\begin{equation}
    d_{PA}=|P(O_P)-P(O_A)|
\end{equation}
\begin{equation}
    d_{HA}=|P(O_H)-P(O_A)|
\end{equation}

Several cases are possible for a given input instance based on the predicted and the actual outcomes and their computed probabilities. The cases may be classified qualitatively in terms of relevancy of the outputs provided. The decreasing order of relevance we consider for different cases is summarized in Table~\ref{table2}. The ordering is based on the consequence of individual cases as explained later. Thus, \emph{ErrScore} is a score that is obtained by quantifying these cases.

\begin{table}[!t]
\renewcommand{\arraystretch}{1.3}
\caption{Qualitative Relevance}\label{table2}
\begin{center}
     \begin{tabular}{ | c | c | c | c |}
     \hline
     CASE & OUTCOME & PROBABILITIES & QUALITATIVE \\
          &         &               & RELEVANCE \\
     \hline
     1 & $O_P = O_A$ & - & Highly Relevant \\ \hline
     2 & $O_P \neq O_A$ & $P(O_H)=P(O_P);P(O_P)>P(O_A)$  & Moderately Relevant \\ \hline
     3 & $O_P \neq O_A$ & $P(O_H)>P(O_P)>P(O_A)$ & Relevant \\ \hline
     4 & $O_P \neq O_A$ & $P(O_H)>P(O_A)>P(O_P)$ & Less Relevant  \\ \hline
     5 & $O_P \neq O_A$ & $P(O_H)=P(O_A); P(O_A)>P(O_P)$ &  Irrelevant   \\
     \hline
     \end{tabular}
 \end{center}
 \end{table}

Case 1: $O_{P}=O_{A}$

In this case, for a given input instance in the test set, the predicted outcome and the actual outcome are equal. Thus, there is no error in the predicted outcome.

Case 2: $O_P\neq O_A; P(O_H)=P(O_P)>P(O_A)$

In this case, for a given input instance the predicted outcome and the highest probable outcome are equal but a different output (actual outcome) is selected. Since the predicted outcome is equal to the highest probable outcome $(d_{PH}=0)$ and the classification model has not been able to capture the switch in output, the error is kept minimum and is equal to $\beta\cdot d_{PA}$  where $\beta$ is a positive real constant, whose value depends on the application.

Case 3: $O_P\neq O_A; P(O_H)>P(O_P)>P(O_A)$

In this case, for a given input instance the predicted outcome, the actual outcome and the highest probable outcome are not equal. Since the probability of the predicted outcome lies in between that of the highest probable outcome and the actual outcome, the error value is higher than the previous case, and is equal to $(\alpha\cdot d_{HP}+\beta\cdot d_{PA})$ where $\alpha$  and $\beta$ denote positive real constants.

Case 4: $O_P\neq O_A; P(O_H)>P(O_A)>P(O_P)$

In this case, for a given input instance the predicted outcome, the actual outcome and the highest probable outcome are not equal. Since the probability of the predicted outcome lies farther away than that of the highest probable outcome and the actual outcome, the error value is much higher, and equal to $(\alpha\cdot d_{HP}+\beta\cdot d_{PA})$ as in Case 3.

Case 5: $O_P\neq O_A; P(O_H)=P(O_A)>P(O_P)$

In this case, for a given input instance the actual outcome and the highest probable outcome are equal but the classification model selects a different output (predicted outcome). Since the actual outcome is equal to the highest probable outcome and the performance of the classification model was poor to select a different output, the error rate is much higher than that of the previous cases. The error is equal to $(\alpha+\beta)\cdot d_{HP}$.

Combining above equations and normalizing over $(\alpha+\beta)$, we find the following error score function

\begin{equation}
    ErrScore=\frac{\alpha(d_{HP})+\beta(d_{PA})}{\alpha+\beta}.
\end{equation}

Therefore, the \emph{RS} for a sample is computed as below

\begin{equation}
    Score=(1-ErrScore)\times 100 ,
\end{equation}

\begin{equation}
    Score=(1-\frac{\alpha(d_{HP})+\beta(d_{PA})}{\alpha+\beta})\times 100 .
\end{equation}

\section{Experimentation and Results}\label{ER}
In this section, we study the performance of the proposed \emph{RS} metric and compare it to the \emph{CA} metric for different classifiers used in the breakout dataset. We further investigate the influence of $\alpha$  and $\beta$ parameters of the RS metric function.

The performance of \emph{RS} metric is studied through various experimentations using the Weka~\cite{WK} simulator on the following rule-based prediction models: DecisionTable~\cite{RK}, JRip~\cite{WC}, Nearest Neighbor with generalization (NNge)~\cite{BM}, PART~\cite{EF}, ConjunctiveRule~\cite{PC} and Ridor~\cite{BG} on the breakout dataset~\cite{AG}.  The breakout dataset consists of 236 samples of data from different users gathered from the breakout area. Here 70\% of the dataset is used for training and the remaining 30\% is used as test set. We consider 10 randomly shuffled training and test sets for experimentation and then mean the results to avoid biased results. We perform the following investigations:

1.	Comparison of \emph{RS} and \emph{CA} Metrics

2.	Significance of $\alpha$  and $\beta$

3.	Lower and Upper Bounds of RS with changing $\alpha$ and $\beta$

4.	Testing of \emph{RS} on Random Output Data

\subsection{Comparison of RS and CA Metrics}
\begin{figure}[!t]
    \centering
        \includegraphics[height=8cm]{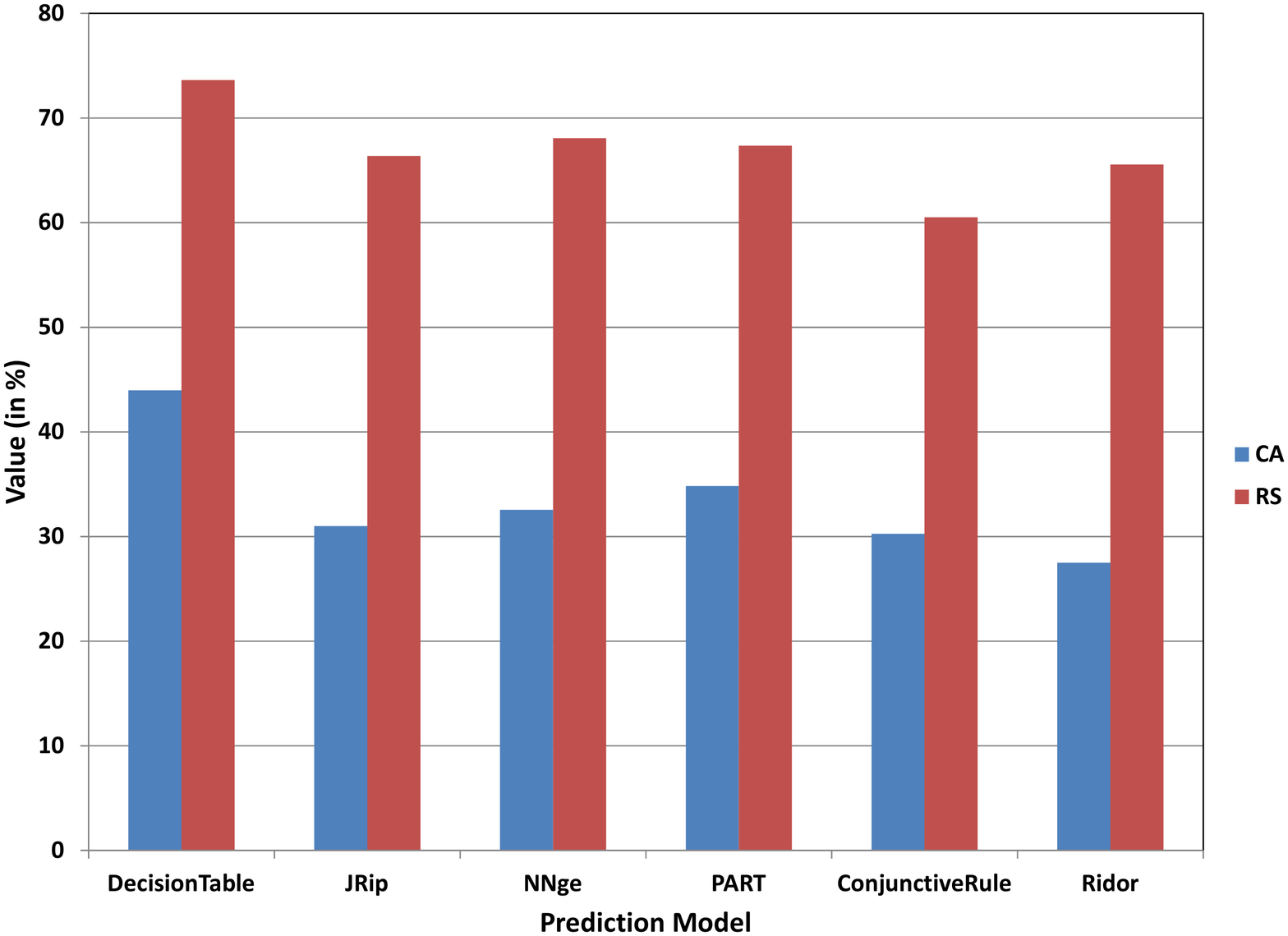}
    \caption{Performance Metric vs. Prediction Model}\label{fig3}
\end{figure}

  Fig.~\ref{fig3} shows the graph of Performance Metric vs. Prediction Model for two different metrics: \emph{CA} and \emph{RS}. The RS values are computed with $\alpha=2$ and $\beta=1$. As discussed earlier, low accuracy values are achieved with the \emph{CA} metric. This means that no prediction models provide exact lighting conditions based on the input instances in more than 50\% of the test samples. However, with RS as a metric, a maximum of 73\% relevant lighting condition is achieved with DecisionTable prediction model.

  An interesting observation is that the prediction model Ridor has lower \emph{CA} than Conjunctive Rule, whereas Ridor has a better \emph{RS} value than Conjunctive Rule. This is interpreted as follows. For a portion of the data samples, the most desired outcome (the best match) is calculated for each prediction model and ConjunctiveRule performs better than Ridor, i.e. it has higher classification accuracy. However, when it comes to the relevance of the predicted outcomes, the Ridor is more successful than ConjunctiveRule, i.e. it predicts the most relevant outcomes on average. Even though ConjunctiveRule finds the best match slightly more often than Ridor, when it cannot find the best match, it comes up with less relevant predictions.  Therefore, Ridor is preferable for applications where relevance of the predicted outcome is more critical than accuracy. As a consequence, for such applications, \emph{RS} as a performance metric is more relevant.

\subsection{Significance of $\alpha$  and $\beta$}
\begin{figure}[!t]
    \centering
        \includegraphics[height=8cm]{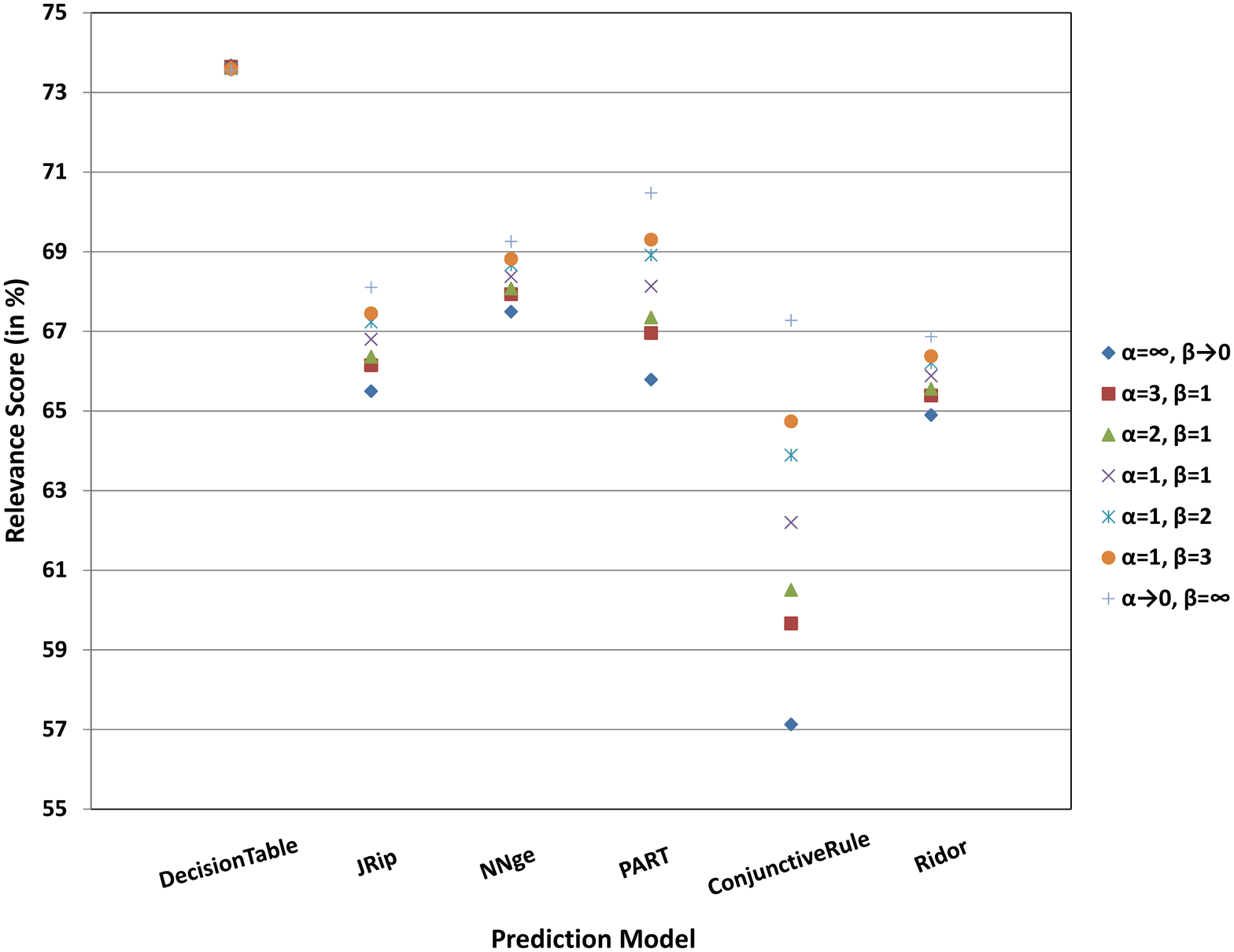}
    \caption{Relevance Score vs. Prediction Model}\label{fig4}
\end{figure}

In this experiment, we investigate the influence of varying values of $\alpha$  and $\beta$ on \emph{RS}. Fig.~\ref{fig4} shows the plot of \emph{RS} vs. Prediction Model for different values of $\alpha$ and $\beta$. The graph shows that the prediction model DecisionTable is more consistent with the changing ($\alpha$, $\beta$) whereas the \emph{RS} values improve with $\alpha$  decreasing and $\beta$ increasing for other prediction models.  The parameters $\alpha$  and $\beta$ are used as weights for probabilistic distances $d_{HP}$ and $d_{PA}$, respectively. Thus, when $\alpha$  is higher the product $\alpha \cdot(d_{HP}$) is emphasized more, i.e. the probabilistic distance between the predicted and actual outcome is given less importance. When $\beta$ is higher, the product $\beta\cdot(d_{PA})$ is emphasized more, i.e. the probabilistic distance between the most probable outcome and the predicted outcome is given less importance.

\begin{figure}[!t]
    \centering
        \includegraphics[height=8cm]{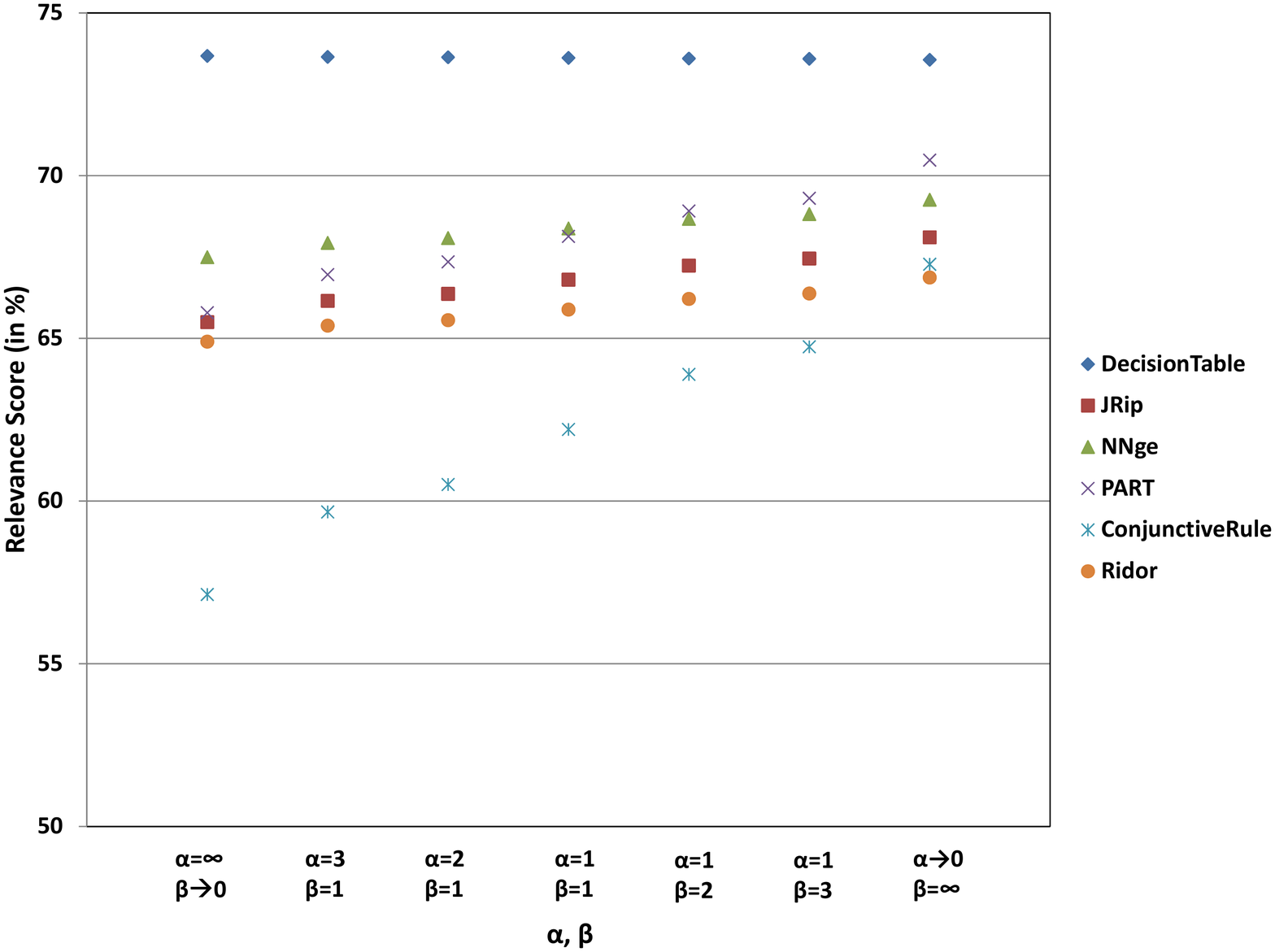}
    \caption{Relevance Score vs. ($\alpha$, $\beta$)}\label{fig5}
\end{figure}

Fig.~\ref{fig5} shows the plot of \emph{RS} vs. $\alpha$, $\beta$ for different prediction models. From the graph, we find that the predictions made by ConjunctiveRule model are closer to the actual outcome and hence varying $\alpha$, $\beta$ has more influence on the RS values. The low RS values for the ConjunctiveRule are due to the fact that the model is not good in consistently predicting relevant outcomes for the given scenarios.

\subsection{Lower and Upper Bounds of RS with changing $\alpha$ and $\beta$}
Maximum or minimum \emph{RS} that a prediction model can reach is found by taking either $\alpha$ or $\beta$ to be very large.

When $\alpha$ is very large compared to $\beta$, the error is computed is as follows:

\begin{equation}
    ErrScore=\lim_{\alpha\to\infty} \frac{\alpha(d_{HP})+\beta(d_{PA})}{\alpha+\beta}=\lim_{\alpha\to\infty} \frac{\alpha(d_{HP})}{\alpha}=d_{HP} .
\end{equation}

When $\beta$ is very large compared to $\alpha$, the error is computed is as follows:

\begin{equation}
     ErrScore=\lim_{\beta\to\infty} \frac{\alpha(d_{HP})+\beta(d_{PA})}{\alpha+\beta}=\lim_{\beta\to\infty} \frac{\beta(d_{PA})}{\beta}=d_{PA} .
\end{equation}

The lower and the upper bounds on RS for the considered prediction models are summarized in Table~\ref{table3}.

\begin{table}
\renewcommand{\arraystretch}{1.3}
\caption{Summary of maximum and minimum \emph{RS} values for various prediction models on the breakout dataset}\label{table3}
\begin{center}
     \begin{tabular}{ | l | c | c |}
     \hline
     & $\alpha\rightarrow \infty$ & $\beta\rightarrow \infty$   \\ \hline
      DecisionTable & 73.68 & 73.56 \\ \hline
      JRip & 65.50 & 68.10 \\ \hline
      NNge & 67.50 & 69.26 \\ \hline
      PART & 65.79 & 70.48  \\ \hline
      ConjunctiveRule & 57.13 &  67.28   \\ \hline
      Ridor & 64.90 & 66.87 \\
     \hline
     \end{tabular}
 \end{center}
 \end{table}

\subsection{Testing of RS on Random Output Data}
\begin{figure}[!t]
    \centering
        \includegraphics[height=8cm]{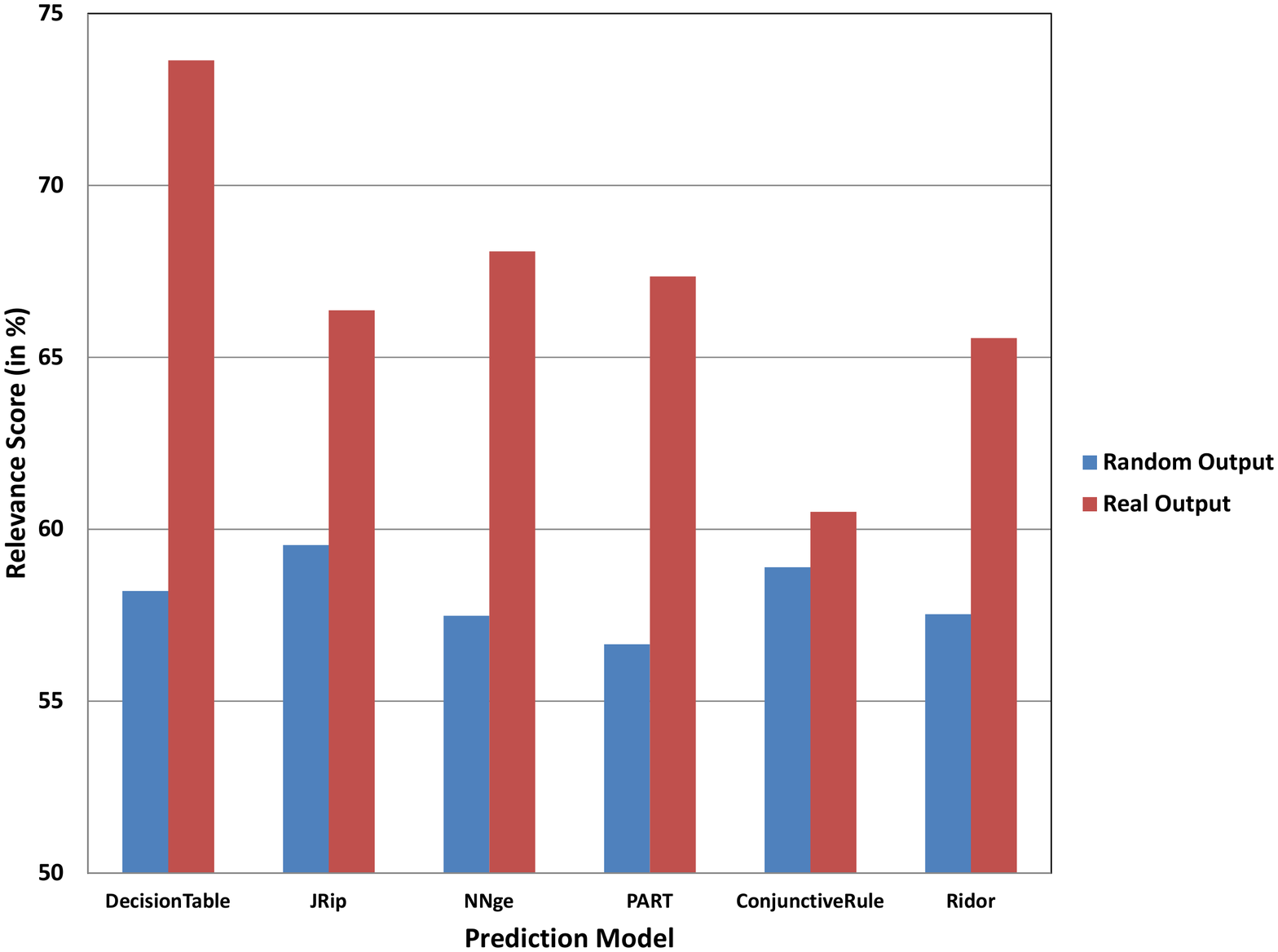}
    \caption{Relevance Score vs. Prediction Model}\label{fig6}
\end{figure}

In this test, we study the \emph{RS} results on the same dataset with randomly generated output. One of the main motivations for this research was the fact that the performance of the prediction model should not be punished for inconsistencies in the data pattern that are caused by factors outside the observed context. For example, a prediction model whose goal is to find the desired lighting conditions for the user can be expected to perform at best as good as the user himself. If the user is generating inconsistent data, i.e. if an observed context maps to different outcomes at different times, then the student (the prediction model) will be learning from an inconsistent teacher (the user), whose reasons for changing mind are completely hidden from the student, i.e. the reasons either don't exist or they are beyond the observed context. Consider the extreme case where there is no relation between the input instances and output lighting conditions, i.e. the user throws an 8-sided dice to select the desired lighting condition and the dice is not part of the observed context. In this case, the CA metric would be equal to 1/8 as there are 8 possible lighting conditions. We also expect to achieve a lower RS in comparison to the scores on the dataset with real output. On the other hand, note that the user makes no distinction between lighting conditions. For a given observed context, even though there is exactly one desired lighting condition, all possible lighting conditions are indeed somewhat relevant. Therefore, we expect that the RS metric will be higher than the CA metric. Fig.~\ref{fig6} shows the plot of \emph{RS} vs. prediction model for random output data and for the real output data from the breakout dataset.

 In the random output dataset, since one of the eight light conditions is chosen at random i.e. with probability 0.125, the output probability distributions are almost uniform for the input instances. This makes it difficult for the prediction models to select the right output choice. Thus all the considered prediction models lie between the range 50 and 60\%. But in case of real output dataset, there are some relationships between the input and output causing the output probability distribution to be non-uniform. Thus the prediction model DecisionTable performs the best with a \emph{RS} of 74\%.

From the performed studies, we find that RS is a more suitable metric than the commonly used metrics for the mentioned class of applications. Furthermore, the RS metric provides a provision to select the prediction models that have different characteristics in predicting the outcomes. This depends on the application requirement, and is done by varying parameters $\alpha$ and $\beta$. For example, a higher value of $\alpha$ is chosen as compared to $\beta$ for applications where capturing inconsistencies in the outcome for an observed context is not so critical.

\section{Conclusion}\label{CONC}
The choice of a suitable performance metric is critical to select a machine learning algorithm for intelligent applications. In this direction, we presented the drawback of the commonly used evaluation metric \emph{CA} with an illustration of intelligent lighting and similar class of applications. We further proposed a new metric named as \emph{Relevance Score} to evaluate machine learning algorithms when there is more than one choice of output class labels suitable for an input instance. The metric is based on the probability values of the outcomes computed from the dataset for a given observed context. We also presented a detailed analysis on the performance and relevance of the proposed metric \emph{RS} through various experiments on the breakout dataset. From the obtained results, we conclude that \emph{RS} is a better performance metric for the class of applications where the relevance of predictions is critical rather than the prediction accuracy. Moreover, variants of RS computation can be used by carefully selecting the parameters of the RS computation function based on the application requirement.

\subsubsection*{Acknowledgments.}
We would like to acknowledge Smart Context-aware Services (SmaCS) project for supporting this work. We would also like to thank Serge Offermans for providing the breakout dataset.

\end{document}